\documentclass[11pt]{article}

\usepackage[final]{acl}

\usepackage{times}
\usepackage{latexsym}

\usepackage[T1]{fontenc}
\usepackage[utf8]{inputenc}

\usepackage{microtype}

\usepackage{inconsolata}

\usepackage{graphicx}
\usepackage{amsmath}
\usepackage{multirow}
\usepackage{algorithm}
\usepackage{algpseudocode}
\usepackage{listings}
\usepackage[table]{xcolor}
\usepackage{enumitem}
\usepackage{tcolorbox}
\usepackage{amssymb}  
\usepackage{booktabs}
\usepackage[utf8]{inputenc}
\title{Enhancing Repository-Level Code Completion with Reinforcement Learning in Real-World Scenarios}

\author {
    Zhenlong Dai\textsuperscript{\rm 1,2}, 
    Zhuoluo Zhao\textsuperscript{\rm 1}, 
    Xiu Tang\textsuperscript{\rm 1}, 
    Sai Wu\textsuperscript{\rm 1}, \\
    \textbf{Chang Yao}\textsuperscript{\rm 1,2}, 
    \textbf{Zhipeng Gao}\textsuperscript{\rm 1}\footnotemark[1], 
    \textbf{Jingyuan Chen}\textsuperscript{\rm 1}\thanks{co-corresponding authors.}
\\
\textsuperscript{\rm 1}Zhejiang University\\
\textsuperscript{\rm 2}Innovation and Management Center,\\ School of Software Technology (Ningbo), Zhejiang University\\
\{zhenlongdai, zhuoluozhao, tangxiu, wusai\}@zju.edu.cn\\
\{changy, zhipeng.gao, jingyuanchen\}@zju.edu.cn
}
\newcommand{\tool}{{IntentCoder}}

\begin{document}
\maketitle
\begin{abstract}
The rapid development of large language models (LLMs) for code, which closely aligns with the practical needs of real-world software development, has sparked growing attention in repository-level code completion.
However, existing Code LLMs present challenges in focusing on suitable contexts from the repository and in deeply reasoning about cross-file dependencies.
To address these challenges, we propose a novel reinforcement learning framework for repository-level code completion.
To better understand both in-file and cross-file contexts, we employ identifier-driven intent recognition to capture completion intent, thereby improving the model's performance in real-world scenarios.
To enhance cross-file reasoning,  we propose a reward-driven completion learning to effectively reward in complex repository completion scenarios. 
To guide LLMs in completing code that aligns with intended functionality and repository context, we introduce a selective-exploration strategy that directs the model to focus on low-confidence, high-reward tokens, promoting the exploration of valuable, underexplored completion patterns.
Experimental results show that our approach significantly improves the performance of Code LLMs.
The code and the dataset are available at \url{https://github.com/zhenlongDai/RepoCompletion}.
\end{abstract}

\begin{figure}[t]
\includegraphics[width=0.98\linewidth]{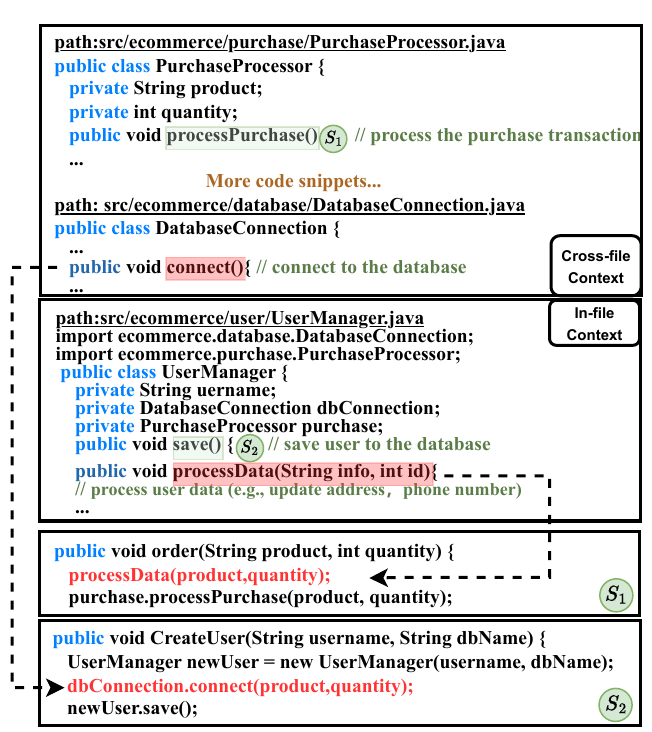}
\vspace{-0.5em}
\caption{Situations of repo-level code completion. 
%Situation (1) reveals that the model gives insufficient consideration to cross-file context and overly focuses on in-file context, which results in incorrect completions. Situation (2) demonstrates that the model’s excessive focus on cross-file context also leads to errors.
}
\label{fig:introduction}
\vspace{-1.0em}
\end{figure}

\section{Introduction}
With the rapid development of large language models (LLMs) for code~\cite{jiang2024survey,zheng2023survey}, these LLMs have demonstrated advanced capabilities in code understanding and completion. 
Many LLMs are integrated into IDEs to support developers' daily development, one of the most common usage scenarios is code completion~\cite{zhou2022improving,wang2023practitioners}.
Previous studies have explored code completion tasks on different granularities, such as line-level code completion~\cite{liu2024non,izadi2022codefill}, function-level code completion~\cite{li2022competition,quan2025codeelo}. 
More recently, the repository-level code completion~\cite{zhang2024codeagent} has become increasingly important because it better aligns with the practical needs of developers in real-world software development. 
However, repo-level code completion is a non-trivial task by considering the following situations: 

As shown in Fig.~\ref{fig:introduction} (s1), this example shows the LLM incorrectly focusing on method signatures in the local (i.e., in-file) context, while ignoring the broader global (i.e., cross-file) context.  
The LLM incorrectly invokes \texttt{processData(product, quantity)}, which fails to reflect the intended action of placing an order. 
The proper call is \texttt{processPurchase(product,quantity)} (\textit{e.g.}, colored in green), which aligns with the intended purchase logic defined in the \textit{PurchaseProcessor} class, imported from the cross-file context. 
\textbf{This situation highlights that the LLM can't correctly focus on local context or global context.}
As shown in Fig.~\ref{fig:introduction} (s2), this example shows \textbf{an LLM's failure to deeply reason about the target cross-file object.}  
The LLM incorrectly calls \texttt{dbConnection.connect()} directly after instantiating \textit{newUser} neglecting the initialization logic \texttt{newUser.save()} encapsulated within the \textit{UserManager} class, which internally manages the database interaction. 
%\textbf{This situation demonstrates the second challenge of repository-level code completion, i.e., how to assist LLMs to deeply reason about cross-file context.} 

Therefore, a key research question regarding repo-level code completion stands out, i.e., \textit{how to improve the model's capability to utilize both cross-file and in-file context effectively?}
It is a challenging task:
(i) \textbf{Guiding LLMs to focus on the suitable context (e.g., in-file context or cross-file context or both) remains a non-trivial problem.}
Different parts of the codebase depend on vastly different types of contextual information. 
Some completions can be resolved using local (in-file) scope, while others require a broader, cross-file perspective. 
LLMs still lack the capability to decide which type of context is relevant for a given completion task~\cite{ding2023crosscodeeval}. 
% Put in motivation: This difficulty arises from their excessive sensitivity to context, which leads to two main issues: on one hand, the model overly relies on the entire repository's context, mistakenly assuming that cross-file information is always necessary; on the other hand, it may focus too narrowly on the current file, ultimately resulting in completion failures.
% As a result, the model is unable to adapt to the real-world repository-level programming environments.
(ii) \textbf{LLMs still struggle to reason deeply about the cross-file information due to complicated project dependencies and code structures.} 
From the repository perspective, code completion requires the LLM to have a global view that spans multiple files and modules. 
This means a deep understanding of how different components are architected and interact. 
%For example, completing a method may depend on user-defined types in another file or should be derived from test files located elsewhere in the project. 

To tackle the above challenges, we propose a novel reinforcement learning (RL) approach for repo-level code completion, which is designed to improve the model's completion ability in different completion scenarios.
To address the challenge of focusing on a suitable context, we leverage identifier-driven intent recognition to capture the completion intent.
Specifically, we adopt the GRPO~\cite{shao2024deepseekmath} method for autonomous learning and design an intent reward based on the matching degree between identifiers in the predicted and real code, guiding the LLM to focus on the correct context.
To address the challenges of dependency resolution and cross-file reasoning, we propose a reward-driven completion learning utilizing a reward function for repo-level code completion, providing continuous and granular reward values based on the distance of the code editing as feedback for cross-file reasoning.
%to improve cross-file reasoning for repository-level code completion. Specifically, we
To guide LLMs to effectively complete code that meets the intended functionality and adheres to the repository-level context, we design a selective-exploration strategy optimization that focuses on low-confidence, high-reward tokens in the model's multiple responses, reflecting divergent thought paths and motivating the model to explore underexplored yet more valuable code completion patterns.

In summary, our paper makes the following contributions:
(1) A reinforcement learning method for repo-level code completion, which effectively addresses the issue of rewards in complex repository completion scenarios and improves cross-file reasoning, enabling the model to learn autonomously on real-world repository data.
(2) The first deep research into LLMs' multi-scenario code completion ability, enhancing the model's ability to focus on a suitable repository context through intent recognition and selective exploration.
(3) \tool~achieves state-of-the-art (SOTA) performance on repo-level code completion tasks with code LLM (7B parameters) on RepoBench, demonstrating strong and stable improvements.

\section{Related Work}
\paragraph{Large Language Models for Code.}
A wide range of LLMs~\cite{fried2022incoder, chowdhery2023palm, wang2023codet5+, zheng2023codegeex, christopoulou2022pangu} tailored for code-related tasks, such as code comprehension and generation~\cite{dai2024mpcoder,dai2025less}, have been developed and widely adopted in software engineering to enhance development efficiency~\cite{chen2021evaluating, liu2023repobench}. 
Some open-source models, Deepseek-Coder~\cite{guo2025deepseek} and Qwen2.5-Coder~\cite{hui2024qwen2,dai2026learner}, show strong code generation abilities.
CodeLLMs have shown strong performance on benchmarks~\cite{lai2023ds,allamanis2013mining,lu2021codexglue,raychev2016probabilistic,yu2024humaneval} for single-file code completion, where they can predict whole lines of code effectively. 
However, their performance in repo-level code completion remains limited, as LLMs struggle to capture the full repository context due to token size restrictions~\cite{phan2024repohyper}, focusing mainly on local context within individual files and neglecting cross-file dependencies~\cite{liang2024repofuse}.
% However, their performance in repository-level code completion remains limited.  
% Due to token size restrictions, LLMs struggle to capture the full context of an entire repository~\cite{phan2024repohyper}. 
% Most models are trained  to focus local context within individual files, often neglecting cross-file dependencies in repository-level codebases~\cite{liang2024repofuse}.

\paragraph{Repository-level Code Completion.}
Repository-level code completion aims to provide code completion by leveraging the context of a repository. Given the large size of repositories and model context limitations, many approaches adopt the \textit{retrieve-then-generate} paradigm~\cite{agrawal2023monitor, cheng2024dataflow}. 
Methods like A3-CodGen~\cite{liao2023context}, CodePlan~\cite{bairi2024codeplan}, and RepoFuse~\cite{liang2024repofuse} use static code analysis for contextual integration to improve code completion. 
RepoCoder~\cite{zhang2023repocoder} refines retrieval using feedback from generated code, while RLCoder~\cite{wang2024rlcoder} improves it with a stop signal mechanism.
%CodeAgent~\cite{zhang2024codeagent} and ToolGen~\cite{wang2024teaching} employ agent-based systems for automated completion. 
Despite advancements in context retrieval, effectively utilizing it remains challenging. FT2Ra~\cite{guo2024ft2ra} and CoCoMIC~\cite{ding2022cocomic} integrate retrieved content, but require architectural changes, making them unsuitable for pre-trained LLMs. In contrast, aixcoder-v2~\cite{li2025aixcoder} uses the CoLT method to assimilate long contexts for improved code completion. Unlike previous methods, we use online reinforcement learning to enhance the LLM's performance in various scenarios.

\section{Methodology}

\subsection{Task Definition and Motivation}
\vspace{-0.5em}
\label{task_defintion}
Given an incomplete code snippet $x$ (i.e., in-file context $C_{\text{in-file}}$) and its cross-file contexts $C_{\text{cross}}$ in a specific repository, the task of repository-level code completion is to generate the completion $y$. 
More formally, the task's objective is to generate the code completion $y$ by a model $\pi_{\theta}$ such that the conditional probability $P_{\theta}(y |  C_{\text{in-file}}, C_{\text{cross}})$ is maximized.
The definitions of in-file and cross-file contexts are in Appendix~\ref{sec:Definition}, and the task motivation is in Appendix~\ref{sec:Motivation}.
%, considering both the in-file context and the cross-file context.

\begin{figure*}[thb!]
\centering
\includegraphics[width=0.85\textwidth]{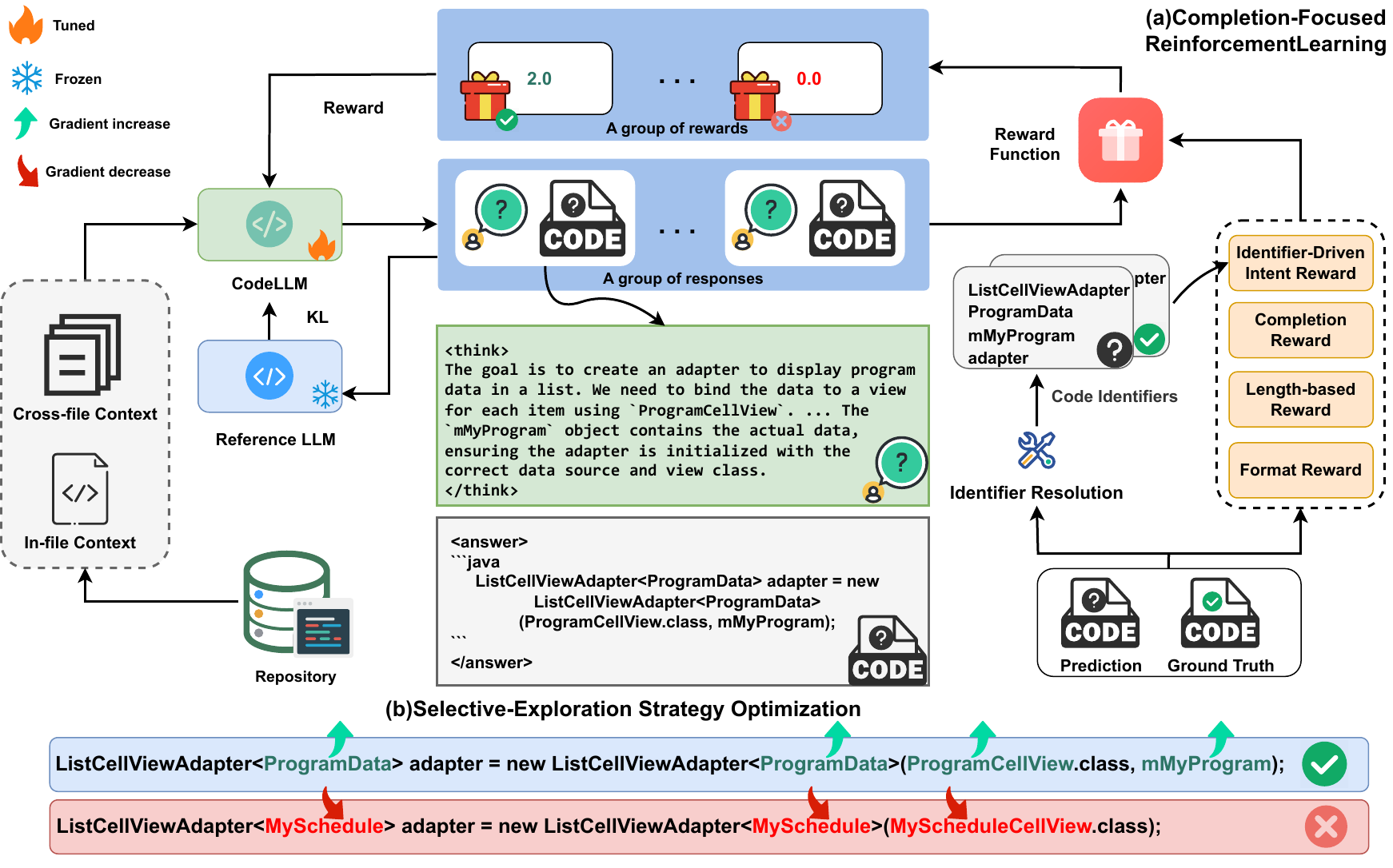}
\vspace{-1.0em}
\caption{Overview of \tool. (a) Illustration of the Completion-Focused RL process. (b) Illustration of the Selective-Exploration Strategy Optimization.
} 
\label{fig:model}
\vspace{-1.0em}
\end{figure*}

\subsection{Preliminary}
\vspace{-0.1em}
\label{GRPO}
Group Relative Policy Optimization (GRPO) is a method of reinforcement learning that explores the potential of LLMs to develop reasoning capabilities, focusing on their self-evolution through a pure RL process~\cite{guo2025deepseek, shao2024deepseekmath}. GRPO is applied to estimate the advantages of language model generations by comparing responses within a query-specific group. 

Specifically, in repo-level completion task, given an input $q = (C_{\text{in-file}}, C_{\text{cross}})$, GRPO samples a group of outputs $\{ y_1, y_2, \ldots, y_G \}$ from the old policy model $\pi_{\theta_{\text{old}}}$, and
then optimizes the policy model $\pi_{\theta}$ by maximizing the following objective:
\begin{equation}
\small
\begin{split}
\label{eq:GRPO}
&\mathcal{J}_{\text{GRPO}}(\theta) = \mathbb{E}_{q \sim P(Q), \{ y_i \}_{i=1}^G \sim \pi_{\theta_{\text{old}}} (y|q)} \frac{1}{G} \sum_{i=1}^{G} \\
&\min(\frac{\pi_\theta(y_i | q)}{\pi_{\theta_{\text{old}}}(y_i | q)} A_i, \text{clip}(\frac{\pi_\theta(y_i | q)}{\pi_{\theta_{\text{old}}}(y_i | q)}, 1  - \epsilon, 1 + \epsilon ) A_i) \\
& - \beta \mathcal{D}_{KL} ( \pi_\theta \| \pi_{\text{ref}}),
\end{split}
\end{equation}
where \( \epsilon \) and \( \beta \) are hyper-parameters, $\mathcal{D}_{KL}$ is the penalty function of Kullback-Leibler divergence. $\pi_{\text{ref}}$ is the reference model, which is initial by the policy model $\pi_{\theta}$. \( A_i \) is the advantage for the $i$-th response, computed using a group of rewards \( \{ r_1, r_2, \ldots, r_G \} \) corresponding to the outputs within each group:
\begin{equation}
\
\begin{split}
\label{eq:Adv}
A_i = \frac{r_i - \text{mean}(r_1, r_2, \ldots, r_G)}{\text{std}(r_1, r_2, \ldots, r_G)}
\end{split}
\end{equation}

\subsection{Completion-Focused Reinforcement Learning}
\label{Reinforcement_Learn}
Even though LLM receives sufficient repository context information, it often fails to effectively leverage and understand it for accurate completion.
Inspired by reinforcement learning, which helps mitigate the reliance on empirical memory for code completion, we propose completion-focused reinforcement learning based on GRPO (Section~\ref{GRPO}) to enable the model to autonomously learn code completion in a repository.

\paragraph{\textbf{Reward-Driven Completion Learning.}}
\label{Completion_Desgin}
Employing reward feedback based on result matching presents challenges in complex and diverse cross-file completion scenarios,
as the model often fails to receive rewards due to the sparsity of signals when directly matching the ground truth code.
To address this challenge, we propose reward-driven completion learning, which provides more precise and granular reward feedback in the complex repo-level code completion environment, enabling the model to receive effective reward signals that enhance its cross-file reasoning ability.

Specifically, given in-file context $C_{\text{in-file}}$ and the cross-file context $C_{\text{cross}} =\{C_i\}_{i=1}^n$ in repository,  the $i$-th output $y_i$ in a group is sampled by the old policy model $\pi_{\theta_{\text{old}}}$, as described by the following equations:
\begin{equation}
\label{old_gen}
\begin{split}
y_i = \pi_{\theta_{\text{old}}}(C_{\text{in-file}}, C_{\text{cross}}).
\end{split}
\end{equation}
Following the prompt of repository-level completion in Fig.~\ref{fig:prompt} (provided in Appendix~\ref{sec:Prompt}), we obtained a group of generated code $\{y_i\}_{i=1}^G$ through multiple samplings by Formula~\ref{old_gen}. Completion reward functions that evaluate the generated code are defined from the following perspectives:
(1) \textbf{completion reward} provides a relative quality ranking of the generated code completions. The completion reward $R_{c}$ for the $i$-th output $y_i$ is defined as follows:
\begin{equation}
\begin{split}
R_{c} = f_\text{Edit}(y_i, \hat{y}),
\end{split}
\end{equation}
where $\hat{y}$ is the ground truth code, $f_\text{Edit}$ is the Levenshtein distance algorithm~\cite{yujian2007normalized} to compare the character differences (\textit{i.e.}, insertions, deletions, substitutions) between generated code and ground truth code. 
This reward provides fine-grained, continuous feedback (ranging from 0 to 1), ensuring the model receives quality feedback even without exact matches.
(2) \textbf{Format reward} $R_{f}$ ensures the model’s output is evaluable by requiring the response to follow a specific format for reasoning and answers. A regular expression checks that these markers are present and correctly ordered.
(3) \textbf{Length-based reward}: Extra code lines may receive spurious rewards, hindering accurate feedback and reducing training efficiency. To address this, we propose a Length-based reward $R_{L}$  to regulate the number of generated lines for better alignment with code completion needs:
\begin{equation}
\small
R_{L}(y_i) = \begin{cases}
1, & if~len(y_i) = 1 \\
0, & otherwise,
\end{cases}
\end{equation}
where $len$ denotes the function that calculates the number of code lines.
The justification for the reward designs is provided in Appendix~\ref{sec:FL_Reward}.
Based on these reward functions, we define the reward function $R$ as follows:
\begin{equation}
\label{eq:Reward}
\small
R = \alpha_c* R_c + \alpha_f* R_{\text{f}} +\alpha_L* R_L,
\end{equation}
where $\alpha_{*}$ is hyper-parameters. 
As shown in Algorithm~\ref{algorithm1}, we use the reward function $R$ to train the model $\pi_{\theta}$, enabling the model to autonomously learn from repository data.

%\vspace{-1.0em}
\label{Intentn_Desgin}
%\vspace{-1.0em}
\begin{algorithm}
\small
\begin{algorithmic}[1]
\State \textbf{Input:} initial policy model $\pi_{\theta}^{init}$, reward function $R$,  dataset $\mathcal{D}=\{C_{\text{in-file}}, C_{\text{cross}} ,\hat{y}\}$; hyperparameters $\mu$
\State policy model $\pi_{\theta} \gets \pi_{\theta}^{init}$

\State reference model $\pi_{\text{ref}} \gets \pi_{\theta}$
\For{step = 1, \dots, M}
    \State Sample a batch $\mathcal{D}_b$ from $\mathcal{D}$
    \State Update the old policy model $\pi_{\theta_{\text{old}}} \gets \pi_{\theta}$
    \State $q \gets \{C_{\text{in-file}}, C_{\text{cross}}\} \in \mathcal{D}_b$
    \State Sample $\{y_i\}_{i=1}^{G} \sim \pi_{\theta_{\text{old}}}(\cdot | C_{\text{in-file}}, C_{\text{cross}})$ for $q \in \mathcal{D}_b$
    \State $\{R_i\}_{i=1}^{G} \gets R(y_i,\hat{y})$ for $y_i \in \{y_i\}_{i=1}^{G}$ (Equation~\ref{eq:Reward})
    \State Compute $A_{i}$ for $y_i \in \{y_i\}_{i=1}^{G}$ (Equation~\ref{eq:Adv})
    \For{iteration = 1, \dots, $\mu$}
        \State Update  $\pi_{\theta}$ by maximizing $\mathcal{J}_{\text{GRPO}}(\theta)$ (Equation~\ref{eq:GRPO})
    \EndFor
\EndFor
\State \textbf{Output:} $\pi_{\theta}$
\end{algorithmic}
\caption{\small{Completion-Focused Reinforcement Learning}}
\label{algorithm1}
\end{algorithm}

\paragraph{Identifier-Driven Intent Recognition.}

In real-world code completion scenarios, the performance of LLMs is often enhanced by utilizing cross-file context from the repository~\cite{ zhang2024codeagent, eghbali2024hallucinator}.
However, the model struggles to focus on suitable contexts from in-file and cross-file contexts~\cite{ding2023crosscodeeval}, resulting in incorrect code completions.
To address this challenge, we propose Identifier-Driven Intent Recognition, which is designed to guide the LLM in identifying the developer's completion intent and understanding code contexts, thereby correctly focusing on suitable contexts. 

Specifically, as shown in Fig.~\ref{fig:prompt}, we instruct the model's reasoning process in the prompt to focus on the user's intent of code completion (\textit{e.g.}, colored in blue), directing its reasoning towards the intent of completion in the reasoning proces.
Given that identifiers contain semantic information from the code, we utilize an identifier-based reward function $R_I$ to evaluate the degree of correctness in the model's intent recognition.
As shown in Fig.~\ref{fig:model}, we parse the code and extract the identifiers (\textit{e.g.}, variable, function, and class names) from the model's prediction $y$ and ground truth code $\hat{y}$. The identifier-based reward function $R_i$ is calculated as follows:
\begin{align}
\small
R_{I} = F_{1}(& f_\text{id}(\pi_{\theta_{\text{old}}}(C_{\text{in-file}}, C_{\text{cross}})),f_\text{id}(\hat{y})),
\end{align}
where $f_\text{id}$ extracts the sets of identifiers from code. $F_{1}$ is the harmonic mean of the precision and recall metric, which is used to evaluate the degree of intent consistency between the predicted and ground truth code.
This approach enables the model to properly consider contextual information and adjust its focus based on the situation, thereby effectively leveraging the advantages of both in-file and cross-file contexts to improve performance.

To effectively coordinate the intent and completion rewards, we join them as follows:
\begin{align}
\small
R_{ic} &= (1+R_{I})*R_c, \\
R &= \alpha_{ic}* R_{ic}  + \alpha_f* R_{\text{f}} +\alpha_L* R_L,
\end{align}
where $R$ is the overall reward function,
$\alpha_{*}$ is hyper-parameter.
By combining these two reward signals, the model is encouraged to prioritize completing the code in a way that both fulfills the user's intended functionality and adheres to the repo-level context. 
When the edit distance between two generated codes is similar in a group of outputs $\{y_i\}_{i=1}^{G}$, the intent reward acts as a coefficient, allowing for a more accurate reward assignment and reducing the lack of distinction in reward values. 
The justification and further analysis of reward design are provided in Appendix~\ref{sec:ID_Reward}.

\paragraph{Selective-Exploration Strategy Optimization.}
\label{Exploration_Optimization}
The model's multiple responses to the same completion task reflect different thought paths, with some responses aligning in part as a consensus, while others show divergence on key points. 
Rewarding the key information where the model shows uncertainty can help cross-file reasoning and contextual attention in the correct contexts.
As shown in Fig.~\ref{fig:model},  both candidates correctly identify \textit{ListCellViewAdapter} (a class in cross-file context). 
The difference is the generic type (\textit{ProgramData} vs. \textit{MySchedule}), which is defined in different files and requires cross-file reasoning to resolve. This type of token (\textit{e.g.}, colored in green/red) significantly influences the reasoning strategy.
However, result-level reward assigns credit to the entire line with the same reward, even when the correct decision depends on specific tokens. 
During the RL process, some key tokens (\textit{e.g.}, colored in green) in the group may have a lower generation probability but receive rewards higher than the group's average.
Based on the above observations, we propose a selective-exploration strategy optimization that emphasizes the tokens that correspond to competing cross-file hypotheses (\textit{e.g.}, \textit{ProgramData}, \textit{MySchedule}) rather than the shared tokens that are predictable (\textit{e.g.}, \textit{ListCellViewAdapter}).  
The key idea is to encourage the model to focus on the competing tokens that represent the high-value and critical aspects of cross-file reasoning and the solution.

Specifically, for the $t$-th token of the $i$-th output $y_{i,t}$ within the group, the confidence function $P$ and value function $V$ are as follows:
 \begin{equation}
  \small
  \begin{aligned}
  P(y_{i,t}) &= \log \pi_\theta(y_{i,t}|q,y_{<t}) \\
  &\quad - \frac{1}{N}\sum_{j=1}^G\sum_{t=1}^{L_i}
  \log \pi_\theta(y_{i,t}|q,y_{<t}),\\
  V(y_{i,t}) &= A(y_{i,t}) - \frac{1}{G}\sum_{j=1}^G A(y_j),
  \end{aligned}
  \end{equation}
  
% \begin{equation}
% \small
% \begin{split}
% P(y_{i,t}) &= \log \pi_\theta(y_{i,t}|q,y_{<t})- \frac{1}{N}\sum_{j=1}^G\sum_{t=1}^{L_i} \log \pi_\theta(y_{i,t}|q,y_{<t}),\\
% V(y_{i,t}) &= A(y_{i,t}) - \frac{1}{G} \sum_{j=1}^G A(y_j),
% \end{split}
% \end{equation}
where $N$ is the total number of tokens in a group of outputs, $L_i$ is the number of tokens in the $i$-th output. Then the tokens that have high value and low confidence are selected as follows:
\begin{equation}
\small
T_{s} = \{ y_{i,t} \mid P(y_{i,t}) < 0, V(y_{i,t}) > 0 \},
\end{equation}
where $T_{\text{select}}$ is the set of selected tokens. To simplify the expression and enhance readability, we define the function $Adv(y_{i,t})$ as follows:
\begin{equation}
\small
\begin{aligned}
\small
&Adv(\cdot)=\min(\frac{\pi_\theta(y_i | q)}{\pi_{\theta_{\text{old}}}(y_i | q)} A_i, \\
&\text{clip}(\frac{\pi_\theta(y_i | q)}{\pi_{\theta_{\text{old}}}(y_i | q)}, 1  - \epsilon, 1 + \epsilon ) A_i).
\end{aligned}
\end{equation}
Based on the function $Adv$, the training of the model $\pi_\theta$ for the position of token $y_{i,t}$ is defined as maximizing the following objective:
\begin{equation}
\small
\mathcal{J}(y_{i,t}) = \begin{cases} Adv(y_{i,t}) - \beta \mathcal{D}_{KL} ( \pi_\theta \| \pi_{\text{ref}}), ~y_{i,t} \notin T_{s}\\
Adv(y_{i,t}), ~y_{i,t} \in T_{s},
\end{cases}
\end{equation}
where token $y_{i,t} \in T_{\text{selected}}$ removes the KL penalty compared to other tokens.
By applying fine-grained gradient updates to tokens in different areas, we adjust the model's attention away from tokens where it already has high confidence and shift it towards those tokens that have the potential to improve the overall output.
This approach encourages the model to gain a deeper understanding of code completion by focusing on divergent contextual areas and exploring valuable code completion patterns.

\section{Experiments}
\subsection{Dataset}
\label{dataset}
To study repository-level code completion in the context of in-file and cross-file scenarios, we construct the \textbf{ICRepo} dataset (\textbf{I}n-file \& \textbf{C}ross-file \textbf{Repo}sitory-level Completion) for model training. 
We collect 32K repo-level code completion examples from the RepoBench-C and RepoBench-P datasets~\cite{liu2023repobench}, with an equal number of Python and Java examples. We contain three completion scenarios as follows:
\textbf{Cross-File-First (CF-F)},
\textbf{Cross-File-Random (CF-R)},
\textbf{In-File (IF)}.
Overall statistics of the dataset are given in Table~\ref{Table-1}.
The training dataset encompasses a variety of real-world code completion scenarios, ensuring the model is exposed to diverse challenges. By incorporating both in-file and cross-file contexts, the dataset covers practical code completion situations that developers frequently encounter in actual projects.
For the test dataset, we use Repobench-Java and Repobench-Python~\cite{liu2023repobench}.
Further details are provided in Appendix~\ref{sec:dataset}.
\begin{table}[thb!]
\small
\resizebox{0.48\textwidth}{!}{
\centering
\begin{tabular}{clrrrr}
\toprule 
\multirow{2}{*}{\textbf{Language}}&\multirow{2}{*}{\textbf{Sets}} & \multirow{2}{*}{\textbf{Num.}}  & \multicolumn{3}{c}{\textbf{Completion position} } \\
& & &IF& CF-F& CF-R\\
\midrule 
\multirow{3}{*}{Java\&Python}&train & 32,000   & 10,100 &11,300 &10,600\\
&val  & 1,000 & 300 &400 &300\\
&test  & 6,000  & 2,000 &2,000 &2,000\\
\bottomrule
\end{tabular}
}
\vspace{-1.0em}
\caption{Dataset Statistics of ICRepo.}
\vspace{-2.0em}
\label{Table-1}
\end{table}

\subsection{Baselines}
We consider the following open-source Code LLMs for comparison, which are widely used in code completion tasks. To simplify the model names, we have removed the ``instruct'' suffix in the following introductions:
\textbf{StarCoder2}~\cite{lozhkov2024starcoder}:  StarCoder2-15B and StarCoder2-7B.
\textbf{DeepSeek-Coder}~\cite{guo2025deepseek}: Deepseek-Coder-6.7B and DeepSeek-Coder-33B.
\textbf{CodeLlama}~\cite{roziere2023code}: CodeLlama-7B and CodeLlama-34B.
\textbf{Qwen2.5-Coder}~\cite{hui2024qwen2}: Qwen2.5-Coder-7B and Qwen2.5-Coder-33B.
\textbf{aixcoder-7B-v2}~\cite{li2025aixcoder} is currently the SOTA 7B model in repo-level code completion.
Then we compare our method with the following methods in the training model for repo-level code completion:
\textbf{fine-tuning}~\cite{hu2022lora},  \textbf{GRPO}~\cite{guo2025deepseek}, \textbf{CoLT}~\cite{li2025aixcoder}.
Further Details are provided in Appendix~\ref{sec:baseline}.

\subsection{Implementation Details}
We train \tool~based on Deepseek-Coder-6.7B and Qwen2.5-Coder-7B, respectively. For each prompt of training data, the old policy model $\pi_{\theta_{\text{old}}}$ generates 6 outputs in a group using a temperature of 0.7 and top-p sampling with a value of 1.0.
To improve training efficiency, we use LoRA~\cite{hu2022lora}, vllm~\cite{kwon2023efficient}, and deepspeed~\cite{rasley2020deepspeed} for train the model $\pi_\theta$. 
Further details are provided in the  Appendix~\ref{sec:Implementation}.

\subsection{Evaluation Metrics}
Following the previous work~\cite{liu2023repobench, eghbali2024hallucinator,zhang2023repocoder, jiang2024aixcoder}, we report performance in two main categories to evaluate the performance of codeLLMs:
\textbf{Code Match}: To evaluate the accuracy of the predicted code lines, we use \textbf{Exact Match (EM)} and \textbf{Edit Similarity (ES)}~\cite{levenshtein1966binary}.
\textbf{Identifier Match}~\cite{ding2022cocomic,ding2023crosscodeeval}: To evaluate the model’s ability to understand the completion intent and predict the correct APIs, we use the following metrics about Identifier:
\textbf{Exact Match (EM)},\textbf{Precision},\textbf{Recall} and \textbf{F1}.
Further details can be found in Appendix~\ref{sec:metrics}.

\begin{table*}%[thb!]
\small
\centering
\resizebox{1.00\textwidth}{!}{
\begin{tabular}{lcc ccccc ccccc}
\toprule
\multirow{4}{*}{\textbf{Model}} & \multicolumn{6}{c}{\textbf{RepoBench-Java}} & \multicolumn{6}{c}{\textbf{RepoBench-Python}} \\
\cmidrule(lr){2-7}\cmidrule(lr){8-13}
& \multicolumn{2}{c}{\textbf{Code Match}} & \multicolumn{4}{c}{\textbf{Identifier Match}} &  \multicolumn{2}{c}{\textbf{Code Match}} & \multicolumn{4}{c}{\textbf{Identifier Match}} \\
\cmidrule(lr){2-3}\cmidrule(lr){4-7}\cmidrule(lr){8-9}\cmidrule(lr){10-13}
& EM & ES &EM & Precision & Recall & F1 & EM & ES & EM & Precision & Recall & F1 \\
\midrule
StarCoder2-7B & 12.20 &  29.78& 15.37&24.64&28.19&25.09&1.80 & 14.47 &3.83  & 7.72 & 13.25&8.40 \\
CodeLlama-7B & 12.40 & 48.50 & 15.90 & 35.60 & 41.60 & 36.46 & 4.47 & 25.56 & 6.37 & 18.26 & 26.68&18.84 \\
aixcoder-7B-v2 &33.07& 68.97& 38.70& 60.57& 65.97& 61.07& 23.20& 59.94& 29.63& 54.70& 62.84& 55.41\\
\midrule
Starcoder2-15B&12.20&28.59&15.67&24.51&27.77&24.92&1.83&13.70&4.07&7.66&13.05&8.34 \\
CodeLlama-34B&12.13&45.39&15.43&33.59&39.11&34.01&4.40&26.02&6.70&18.97&27.55&19.07
\\
DeepSeek-Coder-33B&26.70& 62.03& 31.53& 53.11& 57.99& 53.39& 14.03& 45.00& 18.57& 37.81& 45.49& 38.30\\
Qwen2.5-Coder-32B&35.07&71.13&41.10&63.12&65.28&62.56&29.40&63.82
&35.40&57.52&61.69&57.43\\
\midrule
%\multicolumn{13}{c}{DeepSeek-Coder-6.7B} \\
%\midrule
DeepSeek-Coder-6.7B & 30.20 &66.28& 35.97&57.67&62.81&57.94& 14.97& 43.87& 19.20& 38.13& 47.20& 38.94\\
{DeepSeek-Coder-6.7B}\textsubscript{fine-tuning} & \underline{38.93}&\underline{74.37}&\underline{44.77}&\underline{66.23}&66.17&	\underline{65.22}&31.27&66.36&37.53&59.44&58.76&57.66\\
{DeepSeek-Coder-6.7B}\textsubscript{GRPO} & 35.47&71.29&41.57&63.95&\underline{68.13}&63.83&\underline{34.70}&\underline{68.82}&\underline{40.93}&\underline{62.87}&\underline{60.64}&\underline{60.39}\\
{DeepSeek-Coder-6.7B}\textsubscript{CoLT} & 	35.23&71.52&41.07&61.09&61.94&61.51&	30.20&67.03&40.20&53.66&59.26&56.32\\
\rowcolor{blue!10}
\tool\textsubscript{DeepSeek} & \textbf{41.60}&\textbf{77.64}&\textbf{47.73}&\textbf{69.74}&\textbf{69.18}&\textbf{68.53}&\textbf{35.27}&\textbf{71.12}&\textbf{42.10}&\textbf{64.41}&\textbf{62.98}&\textbf{62.39}\\
\midrule
%\multicolumn{13}{c}{Qwen2.5-Coder-7B} \\
%\midrule
Qwen2.5-Coder-7B &26.80&62.77&32.70&54.74&57.06&53.81&19.97&52.39&25.20&46.55&52.45&46.51\\
{Qwen2.5-Coder-7B}\textsubscript{fine-tuning} &37.43&74.77&43.50&\underline{65.93}&65.78&64.88&31.53&66.61&37.53&59.59&58.69&57.81\\
{Qwen2.5-Coder-7B}\textsubscript{GRPO} &37.50&73.49&\underline{43.60}&65.87&\underline{67.58}&\underline{65.02}&\underline{35.13}&68.71&\underline{41.20}&\underline{62.21}&\underline{60.10}&59.86\\
{Qwen2.5-Coder-7B}\textsubscript{CoLT} & 	\underline{38.80}&\underline{75.33}&42.40&61.40&62.43&61.91&33.00&\underline{69.88}&40.20&61.29&59.81&\underline{60.54}\\
\rowcolor{blue!10}
\tool\textsubscript{Qwen}&\textbf{42.47}&\textbf{77.52}&\textbf{48.57}&\textbf{69.95}&\textbf{69.60}&\textbf{68.75}&\textbf{35.93}&\textbf{71.58}&\textbf{42.70}&\textbf{65.57}&\textbf{63.80}&\textbf{63.37}\\
\bottomrule
\end{tabular}
}
\vspace{-0.5em}
\caption{Evaluation results on the RepoBench-Java/Python dataset. All results in the table are reported in percentage (\(\%\)). The best method is in boldface, and the second best method is underlined for each metric.}
\label{method_compare}
\vspace{-1.5em}
\end{table*}

\subsection{Experimental Results}
\paragraph{Repo-level Code Completion Evaluation and Method Analysis.}
To evaluate the effectiveness of \tool, we apply our training method to Qwen2.5-Coder-7B and DeepSeek-Coder-6.7B, and then compare it with nine Code LLMs and two strong training methods for LLMs.
Table~\ref{method_compare} shows the experimental results of our model and other baselines.
It is obvious that: 
(1) Regarding the comparison with other models, \tool~outperforms all strong baselines by a large margin in both code match and identifier match, surpassing even larger models.
For example, \tool\textsubscript{Qwen} achieves 42.47\% and 35.93\% in Code EM for Java and Python, surpassing the next best model (\textit{i.e.}, Qwen2.5-Coder-32B) by a significant margin of 6.53\% and 5.87\%.
Similarly, in terms of F1 in Identifier Match, \tool\textsubscript{Qwen} outperforms the next best model (\textit{i.e.}, Qwen2.5-Coder-32B) by a significant margin of 5.97\% and 4.97\%.
Compared to models with the same parameters in Code EM, \tool\textsubscript{Qwen} outperforms the current SOTA model (\textit{i.e.}, aixcoder-7B-v2) by a significant margin of 8.53\% in Java and 12.07\% in Python, which clearly demonstrates our model's code completion capability.
(2) Regarding the comparison with other methods, our approach demonstrates substantial improvements across all evaluation metrics over other training methods on two main strong baselines (\textit{e.g.}, Qwen2.5-Coder-7B, DeepSeek-Coder-6.7B). 
Specifically, \tool \textsubscript{DeepSeek} outperforms DeepSeek-Coder-6.7B by up to 11.40\% and 20.30\% in Code EM for Java and Python. Similarly, \tool\textsubscript{Qwen} outperforms Qwen2.5-Coder-7B by up to 15.67\% and 15.96\% . 
\textbf{Overll, \tool~shows substantial and stable improvements compared to other models and methods.}
Further experiments and method analysis are discussed in Appendix~\ref{sec:experiment_and_analysis}.

\begin{table}[thb!]
\small
\resizebox{0.49\textwidth}{!}{
\centering
\begin{tabular}{lcc ccccc}
\toprule
\multirow{3}{*}{\textbf{Model}} & \multicolumn{2}{c}{\textbf{Code Match}} & \multicolumn{4}{c}{\textbf{Identifier Match}}\\
\cmidrule(lr){2-3}\cmidrule(lr){4-7}
& EM & ES &EM & Precision & Recall & F1\\
\midrule
\rowcolor{blue!10}
\tool&\textbf{35.93}&\textbf{71.58}&\textbf{42.70}&\textbf{65.57}&\textbf{63.80}&\textbf{63.37}\\
w/o IDIR&34.60&69.98&41.27&63.74&63.10&62.05\\
w/o RDCL&34.30&68.97&40.20&62.24&60.87&60.25\\
w/o CFCL&\textcolor{gray}{34.47}&67.73&\textcolor{gray}{40.70}&62.06&59.61&59.45
\\
w/o SESO&34.47&70.43&40.97&64.58&62.99&62.39 \\
w/o All &19.97&52.39&25.20&46.55&52.45&46.51\\
\bottomrule
\end{tabular}
}
\vspace{-0.5em}
\caption{Results of Ablation study on the RepoBench-Python dataset in Qwen2.5-Coder-7b.}
\label{ablation}
\vspace{-1.5em}
\end{table}

\paragraph{Contributions of Each Component.}
In this RQ, we conduct an ablation study on \tool\textsubscript{Qwen} to assess the contribution of different techniques by removing key components (\textit{i.e.}, identifier-driven intent recognition (IDIR), reward-driven completion learning (RDCL), completion-focused RL (CFRL), selective-exploration strategy optimization (SESO)) of our approach separately.
For IDIR and RDCL components, we separately remove their the design of the reward function and retain the basic RL process  (\textit{i.e.}, GRPO). 
Similarly, for CFCL, we remove all reward design (\textit{i.e.}, IDIR, RDCL), retain the intention prompt and adopt Code EM as a substitute for reward.
As illustrated in Table~\ref{ablation}, The experimental results show that:
(1) No matter which component we drop, it consistently hurts the overall performance of \tool~across all metrics, signaling the importance and effectiveness of all four components.
When all the components were removed, the performance of the model significantly declined.
(2) When CFCL is removed, the code match and identifier match metrics, except for the EM metric, show a more significant drop than when removing IDIR and RDCL separately. 
This justifies the importance and necessity of CFCL in effectively coordinating the intent and completion rewards, which both fulfill the user’s intended functionality and adhere to the repository-level context.
Notably, EM is slightly higher than removing any one of them (\textit{e.g.}, colored in gray), since we use the exact code match as an alternative to the reward function.
Similarly, the ablation study on Java is provided in Appendix Table~\ref{ablation-appendix}.

%\textbf{RQ2 Summary: } Removing any key component significantly reduces the performance of \tool. 
%Removing Intent-Completion Focused Reward Design has a significant impact, showing its crucial role in coordinating intent and completion rewards to guide CodeLLM’s cross-file reasoning and code completion.

%\subsection{RQ3: Performance of \tool~in Various Completion Scenarios}

\begin{figure*}[htb]
\centering
\includegraphics[width=0.99\textwidth]{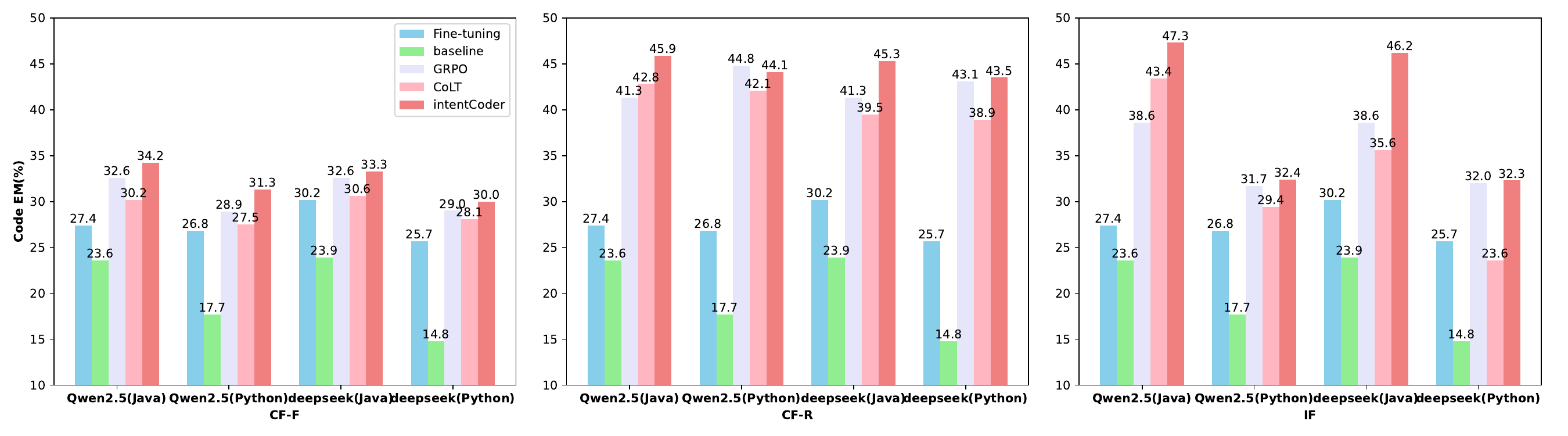}
\vspace{-1.0em}
\caption{Performance of \tool~in Various Completion Scenarios (CF-F,CF-R,IF).
} 
\label{fig:EMval}
\vspace{-1.5em}
\end{figure*}

\paragraph{Robustness Evaluation.}
To evaluate the robustness of our training method, we assess the influence of different training methods on reducing cross-file context-induced failures (\textit{i.e.}, fail) and improving the accuracy of completion (\textit{i.e.}, pass).
As shown in Table~\ref{IC_compare}, compared to the baseline models, our training method significantly reduces cross-file context-induced failures and improves completion accuracy, thereby improving overall performance. 
For instance, our approach has reduced the number of failures of DeepSeek-Coder-6.7b on the Python dataset from -130 to -41 and increased the number of pass from 189 to 709, representing an overall improvement of 668. 
Compared with other methods, our approach reduces the failure rate comparable to GRPO and significantly improves the pass rate compared to the other methods, thereby achieving significant overall improvement.
This justifies the importance and robustness of our method in real-world code completion scenarios, effectively handling and understanding in-file and cross-file contexts. 
%Furher results is provided in Appendix.

% \begin{table}[thb!]
% \setlength{\tabcolsep}{0.2mm}
% \small 
% %\resizebox{0.49\textwidth}{!}{
% \centering
% \begin{tabular}{lrrrrrr}
% \toprule 
% \multirow{2}{*}{\textbf{Method}} & \multicolumn{3}{c}{\textbf{Java}} & \multicolumn{3}{c}{\textbf{Python}} \\
% &fail&pass&overall&fail&pass&overall\\
% \midrule 
% DeepSeek-Coder-6.7B  & -4.37 &+11.7& +7.33&-4.33&+6.3& +1.97 \\
% fine-tuning&-3.9&+19.97&+482&-2.03&+20.3&+18.26\\
% GRPO&\textbf{-78}&+456&+378&\textbf{-40}&+691&+651\\
% \rowcolor{blue!10}\tool\textsubscript{deepseek}&-2.77&\textbf{+21.5}&\textbf{+18.73}&-1.37&\textbf{+23.63}&\textbf{+22.26}\\
% \midrule 
% Qwen2.5-Coder-7B &-128&+301&+173&-138&+225& +87 \\
% fine-tuning&-105&\underline{+597}&+492&-113&+547&+434\\
% GRPO&\textbf{-54}&+548&\underline{+494}&\underline{-62}&\underline{+604}&\underline{+542}\\
% \rowcolor{blue!10}\tool\textsubscript{Qwen}&\underline{-58}&\textbf{+701}&\textbf{+643}&\textbf{-60}&\textbf{+626}&\textbf{+566}\\
% \bottomrule
% \end{tabular}
% %}
% \caption{Result of Code Match EM (\%).}
% \label{IC_compare}
% \end{table}

\begin{table}[thb!]
\small
\centering
\resizebox{0.49\textwidth}{!}{
\begin{tabular}{lrrrrrr}
\toprule 
\multirow{2}{*}{\textbf{Method}} & \multicolumn{3}{c}{\textbf{Java}} & \multicolumn{3}{c}{\textbf{Python}} \\
&fail&pass&overall&fail&pass&overall\\
\midrule 
DeepSeek-Coder-6.7B  & -131 &+351& +220&-130&+189& +59 \\
fine-tuning&-117&\underline{+599}&\underline{+482}&-61&+609&+548\\
GRPO&\textbf{-78}&+456&+378&\textbf{-40}&\underline{+691}&\underline{+651}\\
CoLT&-91&+462&+371&-156&+672&+516\\
\rowcolor{blue!10}\tool\textsubscript{deepseek}&\underline{-83}&\textbf{+645}&\textbf{+562}&\underline{-41}&\textbf{+709}&\textbf{+668}\\
\midrule 
Qwen2.5-Coder-7B &-128&+301&+173&-138&+225& +87 \\
fine-tuning&-105&+597&+492&-113&+547&+434\\
GRPO&\textbf{-54}&+548&+494&\underline{-62}&\underline{+604}&\underline{+542}\\
CoLT&-102&\underline{+635}&\underline{+533}&-125&+603&+478\\
\rowcolor{blue!10}\tool\textsubscript{Qwen}&\underline{-58}&\textbf{+701}&\textbf{+643}&\textbf{-60}&\textbf{+626}&\textbf{+566}\\
\bottomrule
\end{tabular}
}
\vspace{-0.5em}
\caption{Result of Exact Match (EM) count.}
\vspace{-2em}
\label{IC_compare}
\end{table}

\paragraph{Various Scenarios Evaluation.}
To further evaluate the effectiveness of \tool~in various completion scenarios, we compare our approach with other methods (\textit{i.e.}, fine-tuning, GRPO) using the Code EM metric. As shown in Fig.\ref{fig:EMval}, in all completion scenarios (CF-F, CF-R, IF) that are illustrated in Appendix~\ref{sec:scenarios}, our method generally outperforms other methods, although there is one instance where it performs slightly worse. 
The experimental results show that:
(1) In the IF setting, \tool~performs significantly better than other methods, indicating that \tool~is not reliant on guessing based on prior cross-file information and effectively handles in-file contexts.
(2) Similarly, in the CF-F setting, \tool~outperforms other methods consistently, demonstrating that it does not incorrectly depend on in-file context or previous cross-file information, and accurately handles cross-file dependencies, even when encountered for the first time.
Overall, compared to the baseline, \tool~achieves stable improvement and effectively handles different scenarios.

\paragraph{Runtime latency and Case Study.}
To evaluate the runtime latency, We conducted experiments on a test set of 3,000 samples on \tool~using the vLLM inference library. 
% Since the test dataset RepoBench enforces input token limit of 2048, the size of the context fed into our IntentCoder remained below this threshold (on average 1121 tokens), the input latency is under 238 ms. 
RepoBench's max input tokens is  2048, the context fed into IntentCoder averages 1121 tokens, keeping max input latency under 238 ms.
The maximum output token length of the <think> block is 119, with an average length of 20 tokens. The additional latency is small, with an average delay of less than 25 ms and a maximum delay not exceeding 145 ms. 
We believe the small increase in latency is acceptable, as the difference is imperceptible to humans and accuracy is improved.
To illustrate the performance of \tool~compared to other models, we have included a case study in the Appendix~\ref{sec:case_study}.
%\textbf{RQ3 Summary: } \tool~significantly reduces context-induced failures and improves completion accuracy compared to the baseline and other methods, achieving a notable increase in overall. 
%It generally outperforms fine-tuning and GRPO across all completion scenarios, demonstrating \tool's robustness and effectiveness in real-world code completion tasks.

\section{Conclusion}
This research aims to improve the Code LLM's capability of repo-level code completion by enhancing their capabilities in handling both in-file and cross-file contexts. 
To perform this task, we propose a novel training approach designed to improve LLMs' understanding of complex repo-level code by incorporating reward-driven learning for better cross-file reasoning and intent recognition to focus on suitable contexts from in-file and cross-file contexts. 
Selective-exploration strategy optimization guides the model in exploring more valuable code completion patterns, resulting in more accurate and context-aware code suggestions.  
Experimental results demonstrate that the effectiveness of \tool~.  
%We hope that our work advances the improvement of LLMs capable of handling real-world repo-level challenges and provides valuable insights into improving code LLMs for practical use in software development.
We hope our work advances LLMs to handle real-world repo-level challenges and provides valuable insights into improving code LLMs for software engineering development.

\section{Limitations.}
Several limitations are concerned with our work.
Firstly, we do not utilize execution-based rewards due to the complexities involved in constructing executable environments, which are difficult to scale for large datasets. Additionally, the time-intensive nature of software building and testing renders this approach impractical for model training. Although execution-based rewards represent a valuable research avenue, our focus is on rewards derived from edit similarity and identifier matching. This design is more flexible, scalable, and efficient, enabling practical application across large datasets, even when execution is not feasible, and supporting sparse rewards.

Secondly, in this work, we intentionally focused only on strengthening the generation side. i.e., improving code completion results by effectively using cross-file and in-file context. \tool~can naturally be integrated into any retrieval-augmented pipeline as the generator component. However, we believe that research on how to enable LLMs to search and select contexts in a repository within an acceptable time cost and context input limitation is an interesting topic for our future work.

Thirdly, while larger and more capable LLMs may yield better reasoning results during RL training, they come with higher computational costs, such as increased GPU memory usage and longer training times. Due to resource constraints and our goal of supporting community research, we only trained on 7B. We have released our replication package to enable others to replicate our work and verify their ideas.
% Firstly, due to the limited availability of code completion test sets that support code evaluation in other languages, and the difficulty in constructing queries that simultaneously include both intent and incomplete code, our test is based on Python, one of the most popular programming languages used by developers. 
% However, during the training of our method, we used data from two programming languages Java and Python, and we believe that our approach can easily adapt to other programming languages. 
% Secondly, our approach does not explicitly create intent but rather leverages the language comprehension ability of LLMs to reduce the need for learning natural language intent. Exploring how to automatically generate high-quality intent from code is an interesting research topic for our future work.

\section{Acknowledgements}
This research was supported by the National Natural Science Foundation of China (No.62507040), the ``Pioneer'' and ``Leading Goose'' R\&D Program of Zhejiang
(2025C02022),  the Ningbo ``Yongjiang Talent Program'' Youth Innovation Project (2024A-156-G), and Ant Group.

\bibliography{custom}

\appendix

\section{Appendix}
\label{sec:appendix}

\subsection{Definition}
\label{sec:Definition}
In repository-level code completion, in-file contexts and cross-file contexts refer to different types of contextual information that help a model to predict and complete code snippets:

(1) \textbf{In-file Context:} 
Following the previous work~\cite{allamanis2013mining,chen2021evaluating,athiwaratkun2022multi}, in-file contexts refer to the information contained within the prefix of the current file itself, specifically the code that precedes the ground truth code in the incomplete code snippet.
In-file context provides the immediate surrounding code before the incomplete snippet, which typically helps the model understand the structure, syntax, and logic of the code. 
This context is crucial for making predictions when code completion can be inferred solely from the current file's contents, without the need for additional external context.

(2) \textbf{Cross-file Context:}
Cross-file Context refers to the information contained in external files within the repository upon which the current file depends. 
Such information can include functions, classes, or libraries imported into the current file. 
This type of context is extracted by analyzing dependencies or references to other files within the repository~\cite{ding2022cocomic}, providing a broader scope of information for code completion.
Leveraging the cross-file context is essential when the in-file context is insufficient for completing the code. 
%Cross-file context is particularly important when the current file lacks enough information to complete the code, requiring the model to reference external code to make an accurate prediction. 

Both types of In-file and Cross-file contexts are essential for repository-level code completion, allowing the model to make more informed predictions and generate accurate completions.

\begin{table*}[thb!]
\small
\centering
\begin{tabular}{lccrccr}
\toprule 
\multirow{2}{*}{\textbf{model}} &\multicolumn{3}{c}{\textbf{Java}}  & \multicolumn{3}{c}{\textbf{Python}} \\
&In-file&$\rightarrow$&Both&In-file&$\rightarrow$&Both\\
\midrule 
CodeLlama-7B&290&\textcolor{red}{-143}|\textcolor{green}{+227}&374&134&\textcolor{red}{-63}|\textcolor{green}{+100}&174\\
StarCoder2-7B&31&\textcolor{red}{-143}|\textcolor{green}{+192}&80&55&\textcolor{red}{-30}|\textcolor{green}{+29}&54\\
DeepSeek-Coder-6.7B&686&\textcolor{red}{-131}|+\textcolor{green}{351}&906&390&\textcolor{red}{-130}|\textcolor{green}{+189}&449\\
Qwen2.5-Coder-7B&631&\textcolor{red}{-128}|\textcolor{green}{+301}&804&489&\textcolor{red}{-138}|\textcolor{green}{+225}&576\\
DeepSeek-Coder-33B&541&\textcolor{red}{-150}|\textcolor{green}{+410}&801&373&\textcolor{red}{-166}|\textcolor{green}{+214}&421\\
Qwen2.5-Coder-32B&844&\textcolor{red}{-99}|\textcolor{green}{+307}&1,052&771&\textcolor{red}{-126}|\textcolor{green}{+237}&882\\
\bottomrule
\end{tabular}
\caption{Results of Exact Match (EM) Count Code Completion.}
\label{Table_intro}
\end{table*}

\subsection{Motivation}
\label{sec:Motivation}
In this study, we assume that if LLMs can effectively select between global context and local context, the repository-level code completion performance can be significantly boosted.
To further verify our assumption, we conduct a preliminary experiment. 
In detail, we evaluate the performance of utilizing cross-context versus not using it for code completion in the RepoBench datasets (see details about RepoBench in Section~\ref{dataset})
As shown in Table~\ref{Table_intro}, ``In-file'' column refers to correct completion using only the context from the current file. While ``Both'' column denotes correct completions obtained using both in-file and cross-file context. 
During the process of incorporating cross-files, it introduces context-induced failures (\textit{i.e.}, the red numbers). 
For example, CodeLlama-7B shows that the number of correct completions increases from 290 to 374 after using cross-file context, but 143 predictions that were originally correct under in-file context turn incorrect when additional cross-file context is included. 
This reveals a critical challenge: while cross-file context improves overall correctness, it simultaneously destabilizes previously correct predictions. 
It indicates that LLMs still lack reasoning capabilities for cross-file context and motivates us to explore how to guide LLMs to maximize cross-file benefits while mitigating context-induced failures.
The experimental results indicate that while introducing cross-file context improves overall performance, it also harms the model's original performance in in-file scenarios.

\subsection{Justification of Reward Design}
\label{sec:Reward}

\subsubsection{Format and length-based rewards.}
\label{sec:FL_Reward}
The format and length-based rewards are auxiliary mechanisms required to make RL training well-defined for line-level code completion in the repository.
(1) \textbf{Format reward}: The format constraint is a principled modeling choice in GRPO-style RL training. It defines an evaluation protocol so the completion span can be extracted and scored from the model output (e.g., text and code). Without a fixed output format, the reward cannot be computed consistently during training.
(2) \textbf{Length-based reward}: The length-based reward addresses the reward-assignment problem. Our task aims to provide line-level feedback, yet unconstrained generation often produces multiple lines or large segments of a file. As a result, reward triggered by the first correct line propagates through subsequent tokens, causing reward leakage. We observed substantial over-completion in experiments.
Therefore, the length reward localizes the reward to the tokens responsible for the prediction. The strictness is intentional: it prevents reward leakage and leads to more stable training and faster convergence. 
While simple, it effectively ensures that the reward is assigned to the correct position, rather than to unrelated subsequent lines.
In summary, these rewards are important and effective mechanisms to ensure reliable evaluation and proper reward assignment when applying RL to line-level code completion.

\subsubsection{Identifier-Driven Intent Recognition.}
\label{sec:ID_Reward}

\paragraph{Discussion on Identifier Overlap.}
Our method evaluates correctness based on the specific context of repo-level code completion rather than relying on shallow lexical similarity.
We clarify that our task is repo-level code completion, an IDE-style workflow where a developer completes a single line inside an existing file of a repository. 
In this setting, the developer is not freely producing alternative implementations. 
The inserted line must reference concrete program entities available in the current scope and consistent with the surrounding control flow and data flow. 
The identifier appearing in a valid completion is therefore determined by the required binding, such as which object is in scope, which API provides the required return type, and which function is invoked at that program point, rather than by interchangeable naming choice.
Replacing the ground-truth identifier with a different one, even if semantically similar in isolation, usually refers to a different object, breaks the intended call chain, or becomes unusable in the given context. The key issue is whether the completion binds to the intended program entity. 
For this reason, identifier overlap in the intent reward acts as a proxy for correct contextual binding and practical usability, rather than encouraging shallow lexical similarity.

\paragraph{Motivation Behind Using the Intent Reward as a Coefficient.}
 In code completion, correctly capturing the developer’s intent is the primary objective, so we use the intent reward as a coefficient of the completion reward rather than an additive term.
 An additive reward treats intent as an independent bonus and may still assign a high total reward to outputs that are textually similar (\textit{e.g.}, small edit distance) but semantically incorrect. 
 By contrast, using intent as a multiplicative coefficient makes it a gating signal: completions that better match the intended functionality receive proportionally higher reward, while superficially similar but semantically mismatched candidates are down-weighted. 
 This is particularly important when two candidates have similar completion scores due to textual similarity, where the intent signal helps distinguish the one whose reasoning path aligns more closely with the expected behavior.
 From an optimization perspective, the multiplicative form also aligns the gradients of the completion objective with semantic correctness, encouraging the model to prioritize semantically consistent continuations rather than merely minimizing token-level differences.
 
\subsection{Prompt}
\label{sec:Prompt}
As show in Fig.~\ref{fig:prompt}, we use this prompt as the LLMs' input template.
\begin{figure}[hbt!]
\centering
\fcolorbox{black}{gray!6}{%
\parbox{0.90\linewidth}{%
\small
\noindent$\bullet$ \textbf{Instruction}: You are a helpful assistant tasked with completing the next line of a repository-level code file.
 You are provided with an incomplete code file and other code snippets from the same repository. 
 You first think about the reasoning process as an internal monologue \textcolor{blue}{that infers the user's intent of completion}. 
 Then, you should provide the user with the next line of the incomplete code.\\
 Please respond in the following format: \\
 <think>. . .</think>\\
 <answer>$\lstinline|```|$~. . .$\lstinline|```|$</answer>\\
\rule{\linewidth}{0.48pt}
\textbf{Repository name}:$\{$\text{REPOSITORY \textunderscore NAME}$\}$\\
$\bullet$ \textbf{Cross-file context} $C_{\text{cross}}$:\\ 
\text{the path of cross-file context} $C_1$\\
\text{cross-file context} $C_1$ \\
\textbf{. . .}\\
\text{the path of cross-file context} $C_n$\\
\text{cross-file context} $C_n$\\
$\bullet$ \textbf{In-file context} $C_{\text{in-file}}$:\\ 
\text{the path of incomplete code snippet} $x$\\
\text{incomplete code snippet}~$x$
}
}
\caption{the prompt of repository-level completion.} 
\label{fig:prompt}
\end{figure}

\subsection{Dataset}
\label{sec:dataset}

\subsubsection{Completion Scenarios}
\label{sec:scenarios}
We contain three completion scenarios as follows:

\textbf{Cross-File-First (CF-F)}: In this setting, the position of completion is the first appearance of a cross-file line within a file and ensuring there is no identical or similar cross-file invocation for the model to reference. 

\textbf{Cross-File-Random (CF-R)}: In this setting, the position of completion is a random and non-first occurrence of a cross-file line. 

\textbf{In-File (IF)}: In this setting, the position of completion is an in-file line that does not involve any cross-file modules. This setting serves as a robustness test to ensure that the incorporation of cross-file context does not greatly affect the accuracy of predictions.

\subsubsection{Dataset Construction}
We collect 32K repository-level code completion examples from the RepoBench-C and RepoBench-P datasets~\cite{liu2023repobench}, with an equal number of Python and Java examples.
The original RepoBench dataset supports Python$\&$Java with three tasks (Retrieval/Completion/Pipeline). Our dataset is derived from Completion(RepoBench-C) and Pipeline(RepoBench-P) as they directly correspond to in-file and cross-file scenarios. Specifically, the context fed into our IntentCoder includes three parts: Pre-target-line crossfile context, Pre-target-line infile context, and Post-target-line crossfile noise. The Pre-target-line crossfile context refers to import modules that appear before the target prediction line. The pre-target-line infile context refers to code content from the beginning of the current file up to the target line. The post-target-line crossfile noise refers to import modules that occur after the target line, which are regarded as noise since we can only use context before the target line for code completion task. 

For repository-level line-level code completion, cross-file context is obtained by analyzing the classes, functions, identifiers, and other program entities related to the target completion line. 
Leverage their dependency relationships to locate relevant files and the corresponding definitions or functional implementations from the repository, providing repository-level context for completing the target line.

For the test dataset, We use Repobench-Java and Repobench-Python~\cite{liu2023repobench}, each containing 3,000 samples, with 1,000 samples for each of the CF-F, CF-R, and IF settings.
Regarding context size, following the RepoBnech, Our dataset is designed for LLMs with a token limit of 2048. The average context size fed into the model is 1346 tokens for training, 1399 tokens for evaluation, and 1121 tokens for testing.

\subsection{Baselines}
\label{sec:baseline}
We consider the following open-source CodeLLMs for comparison, which are widely used in code completion tasks. To simplify the model names, we have removed the ``instruct'' suffix in the following introductions (\textit{e.g.}, Qwen2.5-coder, CodeLlama, Deepseek-Coder):

\textbf{StarCoder2}~\cite{lozhkov2024starcoder}: The StarCoder2-15B is a 15 billion parameter model trained on 600+ programming languages using 4.3 trillion tokens, while the StarCoder2-7B is a 7 billion parameter model trained on 17 programming languages with 3.7 trillion tokens.

\textbf{DeepSeek-Coder}~\cite{guo2025deepseek}: Deepseek-Coder-6.7B is a 6.7 billion parameter model initialized from Deepseek-Coder-6.7b-base and fine-tuned on 2B tokens of instruction data.
DeepSeek-Coder-33B is the largest variant of the DeepSeek-Coder series.

\textbf{CodeLlama}~\cite{roziere2023code}: CodeLlama-7B and CodeLlama-34B are designed to be safer for use in code assistants and generation applications, with a focus on general code synthesis, understanding, and instruction-following.

\textbf{Qwen2.5-Coder}~\cite{hui2024qwen2}: Qwen2.5-Coder-7B and Qwen2.5-Coder-33B have performed in code generation, reasoning, and fixing, with the latter scaling up training tokens to 5.5 trillion. Qwen2.5-Coder-33B is the SOTA open-source CodeLLM, offering coding abilities comparable to GPT-4o and providing a more comprehensive foundation for real-world applications.

\textbf{aiXcoder-7B-v2}~\cite{li2025aixcoder}: aiXcoder-7B-v2 is currently the SOTA 7B model for repository-level code completion and uses the CoLT method to fine-tune the model to better utilize information within long contexts for code completion tasks.

Then we compare our method with the following methods in the training model for repository-level code completion:

\textbf{fine-tuning}~\cite{hu2022lora}: A supervised fine-tuning (SFT) method where pre-trained models are further trained on a specific task or dataset, refining the model's performance for the target application. We use Low-Rank Adaptation (LoRA) as the specific method.

\textbf{GRPO}~\cite{guo2025deepseek}: Group Relative Policy Optimization (GRPO), which is a method of reinforcement learning. We utilizes reinforcement learning by exact match as a reward to improve code prediction accuracy for repository-level code completion.

\textbf{CoLT}~\cite{li2025aixcoder}: CoLT is a novel training approach for repo-level code completion. It first performs supervised fine-tuning (SFT). After that, preference data is constructed by combining ground truth and model-generated data. Finally, the model is optimized using Direct Preference Optimization (DPO)~\cite{rafailov2023direct}.

\paragraph{Training Setup.}
In our experiment, we carefully control the training setup to make the experimental comparison as fair as possible. 
For the RL method, GRPO and our method use the same training configuration (e.g., the number of epochs, batch size, and rollout size for each prompt). 
CoLT also uses the same training configuration. For Preference Data Construction, we adopt the same count as the rollout size for each prompt to maintain the same data size.
For supervised fine-tuning, we ensure the same overall amount of training data is observed by adjusting the number of epochs, since the RL method samples multiple answers for each prompt. 
The model is trained until convergence, and we report the best checkpoint according to validation performance.
\subsection{Implementation Details}
\label{sec:Implementation}
The rank and alpha values of LoRA~\cite{hu2022lora} are set to 16 and 32, respectively.
We use a learning rate of $1 \times 10^{-4}$ with a linear scheduler and warm-up, with the warm-up ratio set to 0.1.
The model was trained for 1 epoch, consisting of 500 steps, with a batch size of 192.
The model $\pi_\theta$ is trained on 4 NVIDIA A800 (80GB) GPUs for training and 1 GPU for inference simultaneously, which takes 10 hours.
The hyperparameters $\mu$, $\epsilon$, and $\beta$ are set to 1, 0.2, and $1 \times 10^{-3}$, respectively.
The hyperparameters for reward weights, $\alpha_{ic}$, $\alpha_{f}$, and $\alpha_{L}$, are set to 1.0, 0.2, and 0.1, respectively.
We provide well-structured code interface, where reward functions can easily be configured and used. 
Other Details about the implementation are provided in our released code.

During the inference stage, the backbone models we use (e.g., Qwen2.5-Coder and DeepSeek-6.7B-Coder) support substantially longer context windows (up to 128K/16K tokens). Since RepoBench's maximum input tokens are 2048, we calculate the runtime latency based on this setting. 
However, if the cross-file context exceeds 2048 tokens, we can still feed larger contexts to the model and perform reasoning without being constrained by this RepoBench-specific limit.

\subsection{Evaluation Metrics}
\label{sec:metrics}
Following the previous work~\cite{liu2023repobench, eghbali2024hallucinator,zhang2023repocoder, jiang2024aixcoder}, we report performance in two main categories to evaluate the performance of codeLLMs:

\textbf{Code Match}: To evaluate the accuracy of the predicted code lines, we use \textbf{Exact Match (EM)} and \textbf{Edit Similarity (ES)}~\cite{levenshtein1966binary}, which respectively represent the proportion of exact matches and the degree of similarity to the ground truth.

\textbf{Identifier Match}~\cite{ding2022cocomic,ding2023crosscodeeval}: To evaluate the model’s ability to understand the completion intent and predict the correct application programming interfaces (APIs), we use the following metrics:(1) \textbf{Exact Match (EM)}: EM represents the proportion of exact matches to the ground truths in code identifiers.
(2) \textbf{Precision}: Precision measures the proportion of correctly predicted code identifiers out of all predicted identifiers.
(3) \textbf{Recall}: Recall measures the proportion of correctly predicted code identifiers out of all actual code identifiers in the ground truth.
(4) \textbf{F1}: F1 is the harmonic mean of precision and recall metrics, providing a balanced metric for the model's performance.

\begin{table*}[thb!]
%\small
%\setlength{\tabcolsep}{1.0mm}
\centering
\begin{tabular}{lcc ccc ccc}
\toprule
\multirow{4}{*}{\textbf{Model}} & \multicolumn{3}{c}{\textbf{BM25}} & \multicolumn{3}{c}{\textbf{OpenAI text-embedding-ada}} \\
\cmidrule(lr){2-4}\cmidrule(lr){5-7}
& \multicolumn{2}{c}{\textbf{Code}} & \multicolumn{1}{c}{\textbf{Identifier}} &  \multicolumn{2}{c}{\textbf{Code}} & \multicolumn{1}{c}{\textbf{Identifier }} \\
%\cmidrule(lr){2-3}\cmidrule(lr){4-7}\cmidrule(lr){8-9}\cmidrule(lr){10-13}
& EM & ES &EM& EM & ES & EM\\
\midrule
Qwen2.5-Coder-7B &10.33&57.93&18.75&15.71&64.66&26.13\\
{Qwen2.5-Coder-7B}\textsubscript{fine-tuning} &11.87&58.09&19.35&20.35&67.80&27.21\\
{Qwen2.5-Coder-7B}\textsubscript{GRPO} &\underline{13.18}&\underline{60.36}&\underline{20.06}&22.02&68.33&29.41\\
{Qwen2.5-Coder-7B}\textsubscript{CoLT} &12.90&59.72& 19.54&\underline{23.18}&\underline{69.72}&\underline{30.67}\\
\rowcolor{blue!10}
\tool\textsubscript{Qwen} &\textbf{15.33}&\textbf{62.46}&\textbf{22.39}&\textbf{26.79}&\textbf{73.43}&\textbf{38.62}\\
\bottomrule
\end{tabular}
%\vspace{-1.0em}
\caption{Evaluation results on CrossCodeEval-Java.}
\label{CodeEval}
%\vspace{-1.5em}
\end{table*}

\begin{figure*}[thb!]
\centering
\includegraphics[width=\textwidth]{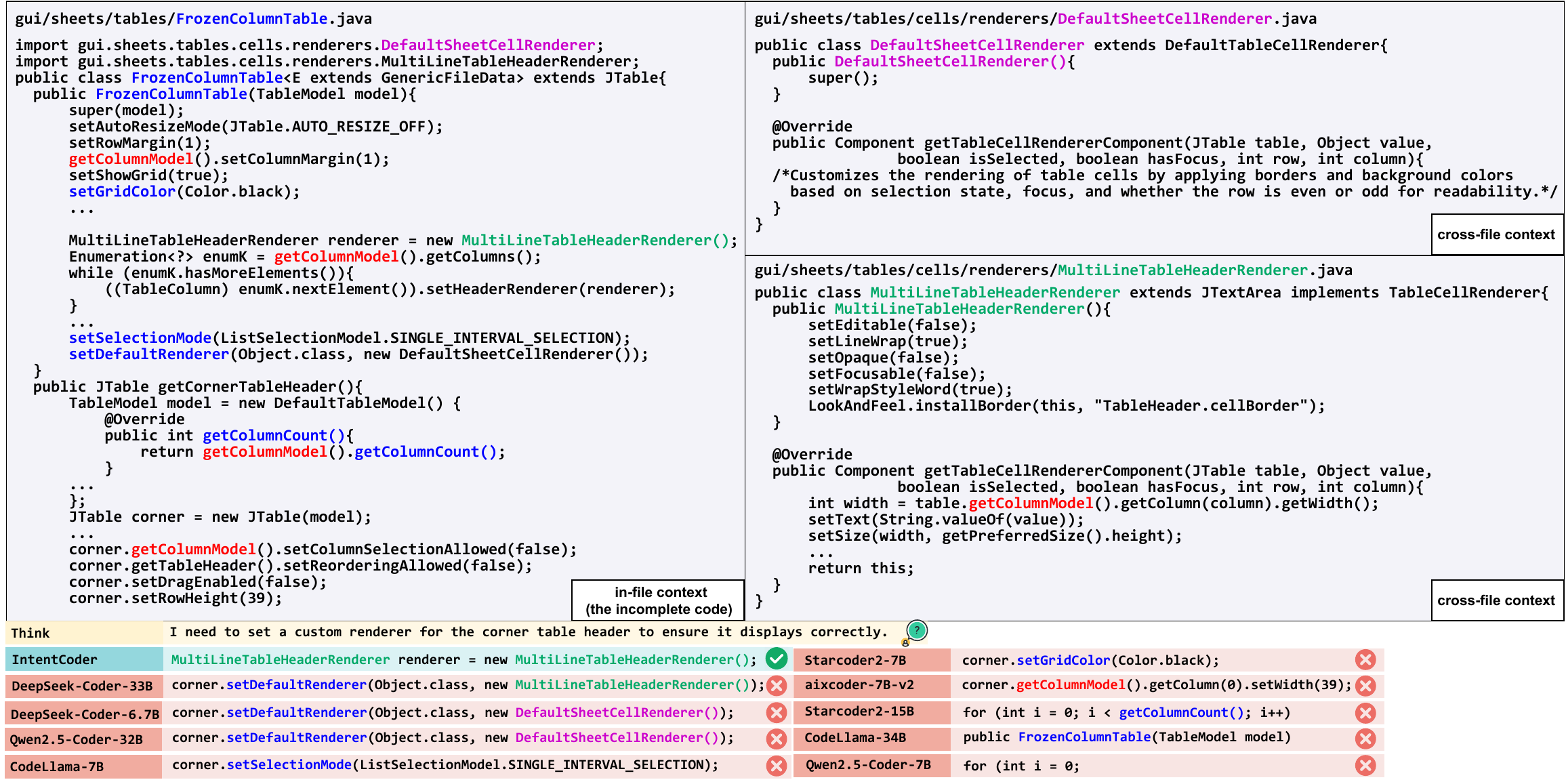}

\caption{The case study for \tool compares it with other Code LLMs.} 
\label{fig:Casestudy}
\end{figure*}

\begin{table*}[thb!]
\small
\setlength{\tabcolsep}{0.2mm}
\centering
\begin{tabular}{lcc ccccc ccccc}
\toprule
\multirow{4}{*}{\textbf{Model}} & \multicolumn{6}{c}{\textbf{RepoBench-Java}} & \multicolumn{6}{c}{\textbf{RepoBench-Python}} \\
\cmidrule(lr){2-7}\cmidrule(lr){8-13}
& \multicolumn{2}{c}{\textbf{Code Match}} & \multicolumn{4}{c}{\textbf{Identifier Match}} &  \multicolumn{2}{c}{\textbf{Code Match}} & \multicolumn{4}{c}{\textbf{Identifier Match}} \\
\cmidrule(lr){2-3}\cmidrule(lr){4-7}\cmidrule(lr){8-9}\cmidrule(lr){10-13}
& EM & ES &EM & Precision & Recall & F1 & EM & ES & EM & Precision & Recall & F1 \\
\midrule
\rowcolor{blue!10}
\textbf{\tool\textsubscript{Qwen}}&\textbf{42.47}&\textbf{77.52}&\textbf{48.57}&\textbf{69.95}&\textbf{69.60}&\textbf{68.75}&\textbf{35.93}&\textbf{71.58}&\textbf{42.70}&\textbf{65.57}&\textbf{63.80}&\textbf{63.37}\\
w/o Identifier-Driven Intent Recognition&40.27&76.77&46.47&68.73&68.62&67.67&34.60&69.98&41.27&63.74&63.10&62.05\\
w/o Reward-driven Completion Learning&40.80&76.94&47.67&69.56&68.49&68.03&34.30&68.97&40.20&62.24&60.87&60.25\\
w/o Completion-focused RL&\textcolor{gray}{40.90}&76.13&47.07&68.49&67.62&66.95&\textcolor{gray}{34.47}&67.73&\textcolor{gray}{40.70}&62.06&59.61&59.45
\\
w/o Selective-Exploration Optimization&40.87&76.75&47.07&68.92&68.41&67.69&34.47&70.43&40.97&64.58&62.99&62.39
 \\
w/o ALL &26.80&62.77&32.70&54.74&57.06&53.81&19.97&52.39&25.20&46.55&52.45&46.51\\
\bottomrule
\end{tabular}
\caption{Results of Ablation study on the RepoBench-java and RepoBench-Python dataset. All results in the table are reported in percentage (\(\%\)).}
\label{ablation-appendix}
\end{table*}

\subsection{Method Analysis and Generalization Experiments}
\label{sec:experiment_and_analysis}

\subsubsection{Method Analysis}
We clarify why our reward method is more appropriate for repo-level code completion task than other methods as follows:

(1) \textbf{Execution-based reward} can provide a strong reward signal, but in repo-level completion, they are difficult to obtain reliably. Our task involves line-level completion in code repositories, where the suffix code context is often missing, so independent compilation is infeasible. Constructing executable environments at scale requires complex dependencies and specific library versions, and building and testing large projects can take minutes or hours, which is impractical for RL training. Thus, while execution-based rewards are valuable, they are not scalable for large cross-file completion datasets. Our rewards are based on edit similarity and identifier matching, which is a more flexible and scalable design. The reward is lightweight and can be applied efficiently across large datasets, ensuring practical use even when execution or validation is unavailable.
In our task, such signals are difficult to obtain reliably: (1) Our task focuses on line-level completion over incomplete files extracted from repositories. These samples typically lack surrounding complete code context or project context (e.g., dependent modules compile ) and therefore cannot be compiled independently, making compiler feedback unstable. (2) Lightweight abstract execution could be applied to a subset of cases, which would substantially reduce coverage and introduce bias toward easily executable examples. For these reasons, we adopt structured static signals (e.g., edit similarity and identifier matching) as a scalable proxy for correctness.

(2) \textbf{SFT reward}: Cross-file completion requires reasoning about how an API is used in a concrete program context. A model trained only with retrieval supervision may learn to associate related files rather than learning to generate correct completions via cross-file reasoning, which may bias the model toward lexical similarity rather than deeper semantic correctness. Our method allows the model to learn such reasoning through reinforcement learning by interacting with the repo-level code-completion environment.

(3) \textbf{Preference-based learning reward}: Preference-based learning methods (e.g., DPO) optimize the model using fixed positive–negative pairs. However, in our setting it is difficult for the model to consistently produce fully correct outputs during self-improvement, and relying on ground-truth pairs alone introduces an out-of-distribution gap between training signals and the model’s own generations. Our reward design instead provides graded feedback: multiple candidate completions are compared and rewarded according to how well they align with program behavior. This allows the model to improve even when none of the candidates is perfectly correct, offering a more stable learning signal than strict binary preference pairs and enabling incremental improvement during training.

The experimental results of fine-tuning and CoLT, when compared with our approach, demonstrate that our approach better facilitates the learning of reasoning abilities in the code repository environment. 
Moreover, the group data generated through model self-distribution can stably and effectively optimize the model's output distribution, thereby obtaining stable and effective learning signals while avoiding unstable signals from ground truth code completion that fall outside the distribution.

\subsubsection{Generalization Experiment.}
We provide further analysis of how the model performs when the retrieved cross-file context contains irrelevant or distracting information outside the RepoBench distribution. 
Specifically, we experimented on the CrossCodeEval-Java~\cite{ding2023crosscodeeval} dataset using different retrievers (BM25~\cite{robertson2009probabilistic} and OpenAI text-embedding-ada-002). CrossCodeEval-Java has 2,139 samples and retrieves the top-5 code snippets for each query. This naturally introduces incomplete and potentially distracting context.
Table ~\ref{CodeEval} shows the experimental results of our model and other methods.
The results show two trends. (1) Under the same retriever, training with our method substantially improves performance (Code EM increases from 10.33\%to 15.33\%), indicating that the model learns to better identify relevant information within a noisy retrieval cross-files context. (2) When retrieval quality improves (e.g., using a more accurate embedding retriever), performance further increases (Code EM increases from 15.33\% to 26.79\%), suggesting that the method can effectively utilize cleaner context even in retrieval noise.
These observations indicate that IntentCoder is not dependent on a particular repository distribution and remains effective under varying levels of irrelevant or incomplete cross-file context.

\subsection{Case Study}
\label{sec:case_study}
We illustrate the effectiveness of \tool~through a case study presented in Fig.~\ref{fig:Casestudy}, In this case, \tool~correctly identifies the user's intent, which is to set a custom renderer for the table corner header (\textit{e.g.}, \texttt{MultiLineTableHeaderRenderer}).  
In contrast, many other models (\textit{e.g.}, DeepSeek-Coder, Qwen2.5-Coder, CodeLlama-7B, Starcoder2-7B, aixcoder-7B-v2), mistakenly think the next step should be related to the corner setup. 
Additionally, due to the lack of intent recognition's capability, many other models overly focus on the in-file context (\textit{e.g.}, colored in blue), such as \texttt{FrozenColumnTable}, \texttt{setGridColor}, \texttt{getColumn}.
At the same time, these models lack cross-file reasoning ability and incorrectly invoke classes from other files (\textit{e.g.}, colored in purple), failing to understand that the appropriate multi-line header renderer should be used.  
By recognizing the overall functional requirements, \tool~accurately provides the correct code completion, avoiding the shortcomings of other models in completion intent recognition and cross-file reasoning.

\end{document}